\documentclass[sigconf,anonymous=False,nonacm,review=false]{acmart}

\AtBeginDocument{%
  }

\copyrightyear{2024}
\begin{document}


\author{Hua Chang Bakker}
\affiliation{
  \institution{University of Amsterdam}
  \city{Amsterdam}
  \country{The~Netherlands}
}
\email{h.c.bakker@uva.nl}

\author{Shashank Gupta}
\orcid{0000-0003-1291-7951}
\affiliation{%
  \institution{University of Amsterdam}
  \city{Amsterdam}
  \country{The~Netherlands}
}
\email{s.gupta2@uva.nl}

\author{Harrie Oosterhuis}
\orcid{0000-0002-0458-9233}
\affiliation{%
  \institution{Radboud University}
  \city{Nijmegen}
  \country{The Netherlands}
}
\email{harrie.oosterhuis@ru.nl}



\begin{abstract}
Variance regularized counterfactual risk minimization (VRCRM) has been proposed as an alternative off-policy learning (OPL) method. VRCRM method uses a lower-bound on the $f$-divergence between the logging policy and the target policy as regularization during learning and was shown to improve performance over existing OPL alternatives on multi-label classification tasks.

In this work, we revisit the original experimental setting of VRCRM and propose to minimize the $f$-divergence directly, instead of optimizing for the lower bound using a $f$-GAN approach.
Surprisingly, we were unable to reproduce the results reported in the original setting.
In response, we propose a novel simpler alternative to f-divergence optimization by minimizing a direct approximation of f-divergence directly, instead of a $f$-GAN based lower bound.
Experiments showed that minimizing the divergence using $f$-GANs did not work as expected, whereas our proposed novel simpler alternative works better empirically.
\end{abstract}

\title{A Simpler Alternative to Variational~Regularized~Counterfactual~Risk~Minimization}
\maketitle

\section{Introduction}
Several methods for off-policy learning have been proposed to learn target policies before deployment \cite{swaminathan2015crm, pmlr-v80-wu18g, dudik2011, gupta-2024-practical, gupta2023safe, Gupta2024, gupta2024optimal, gupta2023deep, gupta2024first}.
One of these methods is variance regularized counterfactual risk minimization (VRCRM) \cite{pmlr-v80-wu18g}, and is based on the minimization of an $f$-divergence between the logging policy and the target policy.
The $f$-divergence is optimized by defining a lower-bound and optimizing the lower bound using an $f$-GAN method \cite{goodfellow2014, nowozin2016}, that is, a generalized generative adversarial network.
In this study, we reproduced and extended the experimental setting of \cite{pmlr-v80-wu18g}, and proposed an alternative way to minimize the divergence via VRCRM.

\section{Existing Approach: VRCRM}
Let $\pi_0$ be the logging policy and let $\pi_\theta$ be a target policy, $\delta(x, a)$ be the loss function for the (context, action) pair $(x,a)$. A standard approach for OPL is via the inverse propensity scoring (IPS) objective~\cite{swaminathan2015crm}:
\begin{equation}
\label{eq:ips}
    \hat R_\text{IPS}(\pi_\theta) = \frac{1}{N} \sum^N_{i=1}  \frac{\pi_\theta(a_i | x_i)}{\pi_0 (a_i | x_i)} \delta(x, a),
\end{equation}
where $N$ is the total number of logged interactions from the logging policy $\pi_0$.
\citet{pmlr-v80-wu18g} defined the $f$-divergence as:
\begin{equation}
\label{eq:div}
    d(\pi_\theta || \pi_0) = \mathbb E_{x \sim \operatorname{Pr}(\mathcal X)} \left[\sum_{a \in \mathcal A} \frac{\pi_\theta(a | x)^2}{\pi_0 (a | x)}\right],
\end{equation}
where $\operatorname{Pr}(\mathcal{X})$ is a probability distribution over the set of context vectors $\mathcal X$. 
If the loss $\delta(x_i, a_i) \leq L$ and divergence $d(\pi_\theta || \pi_0) \leq M$ are bounded, $\phi \in (0, 1)$ and $\lambda = \sqrt{2L \log 1 / \phi}$, then minimizing 
\begin{equation}
\label{eq:prob_upper_bound}
    \hat R_\text{VRCRM}(\pi_\theta) = \hat R_\text{IPS}(\pi_\theta) + \lambda \sqrt{\frac{1}{N} d(\pi_\theta || \pi_0)},
\end{equation}
also minimizes the \emph{true overall loss} of $\pi_\theta$ with probability $1 - \phi$~\cite{pmlr-v80-wu18g}.
The VRCRM method does not optimize the above objective directly, but does so by repeating a two step optimization procedure. 
The first step is to optimize the IPS estimator $\hat R_\text{IPS}$ (Eq.~\ref{eq:ips}) for a single iteration.
The second step is to minimize the divergence using an $f$-GAN~\cite{nowozin2016f}, defined as follows:
\begin{equation}
\label{eq:lower_bound}
    = \sup_{T \in \mathcal{T}} \left\{ \mathbb{E}_{x, y \sim h} T(x, y) - \mathbb{E}_{x, y \sim h_0} f^*(T(x, y)) \right\}, 
\end{equation}
where the function $T$ is modelled using a neural network, and the first term is estimated using the gumbel-softmax trick. The second step is optimized until $d(\pi_\theta || \pi_0) < t/N$, for some threshold $t$ and batch size $N$, or until a predefined number of iterations is exceeded.  
The experiments by \citet{pmlr-v80-wu18g} show that VRCRM is able to reach improved performance on multi-label classification tasks.

\section{A Novel Method Using Simple Approximation}
The use of $f$-GANs to regularize the divergence between the logging policy and the target policy is computationally expensive and based on the minimization of a lower bound on the divergence, losing the theoretical guarantees. 
To avoid optimizing for the $f$-GANs lower-bound, we propose to regularize the divergence using a \emph{direct} sample-based approximation of $d(\pi_\theta || \pi_0)$:
\begin{equation}
\label{eq:div-estimator}
    \hat d(\pi_\theta || \pi_0) = \frac{1}{N} \sum^N_{i=1} \sum_{a \in \mathcal{A}} \frac{\pi_\theta(a | x_i)^2}{\pi_0 (a | x_i)}.
\end{equation}
This approximation thus allows for simpler optimization of Eq. \ref{eq:prob_upper_bound} using the estimator:
\begin{equation}
\label{eq:directly_estimated_obj_actual}
    \hat R_\text{direct}(\pi_\theta) = \hat R_\text{IPS}(\pi_\theta) + \lambda \sqrt{\frac{1}{N} \hat d(\pi_\theta || \pi_0)}.
\end{equation}

\section{Experiments}
All experiments were run ten times and differences in scores were tested for statistical significance using a paired two-sample $t$-test, with significance level $\alpha = 0.05$. The significantly differing scores are marked in the figures as a filled marker. Parameters were as reported by \cite{pmlr-v80-wu18g} or found using validation sets. More details on the experiments are included in Appendix \ref{appendix:params}.

\begin{figure*}
    \centering
    \includegraphics[width=\linewidth]{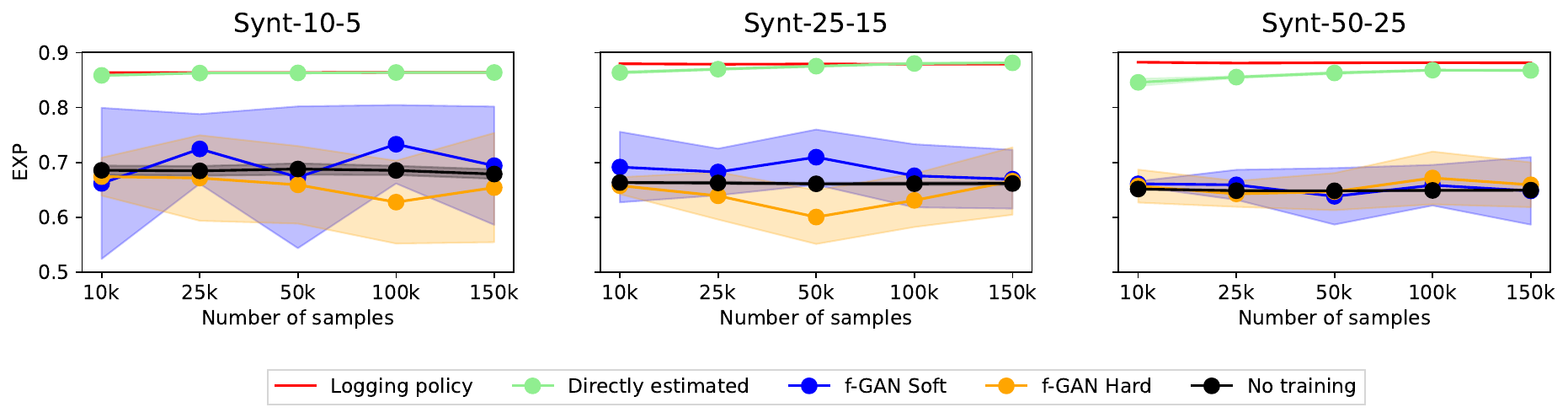}
    \caption{Performance of the methods for divergence minimization. Values closer to the logging policy are better. Scores significantly different from the logging policy scores are indicated with filled markers. Shaded areas indicate standard deviations.}
    \Description{Results of the 1) logging policy, 2) directly estimated divergence, 3) $f$-GAN methods and 4) a neural network without training on three synthetic datasets, which are \texttt{synt-10-5}, \texttt{synt-25-15} and \texttt{synt-50-25}, on dataset sizes varying from 10k samples to 150k samples. The figure shows that the directly estimated is able to match performance with the logging policy, whilst the $f$-GAN policies are unable to.}
    \label{fig:toy_divergence}
\end{figure*}

\subsection{Experimental Setup}
In this work, we adopt the Open Bandit Pipeline (OBP) to simulate, in a reproducible manner, real-world recommendation setups with stochastic rewards, realistic action spaces, and controlled randomization~\cite{obp}.
Synthetic datasets are referred to as \texttt{Synt-a-d}, for example, \texttt{Synt-10-5} denotes a dataset with ten actions and context vectors with five dimensions. 
A reward of one is given if the prediction is correct, otherwise a reward of zero is given.
The training, validation and testing datasets contained the same number of datapoints.

\subsection{Divergence minimization}
\label{section:div_min}
\begin{figure}[tb]
    \centering
    \includegraphics[width=\linewidth]{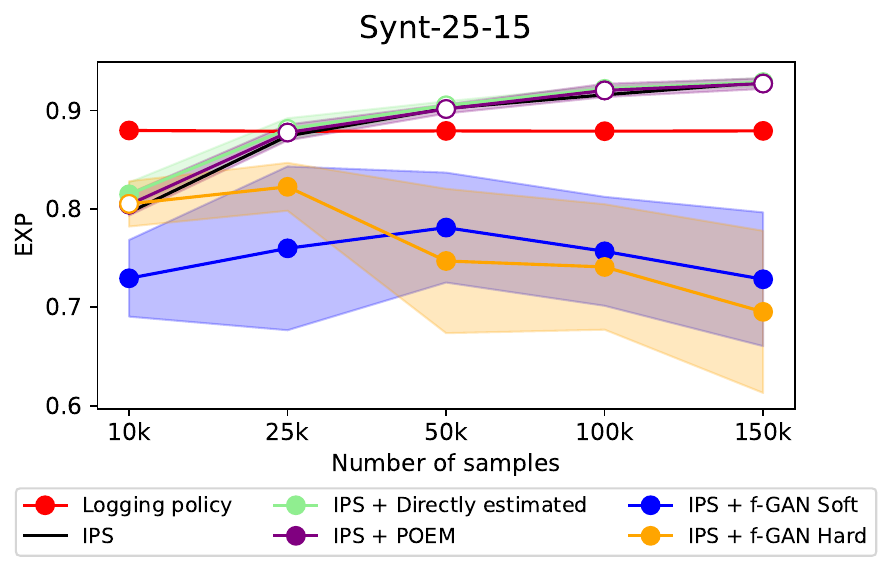}
    \caption{Performance of IPS-based methods on \texttt{synt-25-15}.
    Scores significantly different from the IPS scores are indicated with filled markers. Shaded areas indicate standard deviations.
    }
    \Description{Results of the 1) logging policy and IPS in combination with 2) directly estimated divergence, 3) $f$-GAN methods and 4) POEM on \texttt{synt-25-15}. The figure shows that combining IPS with POEM and directly estimated divergence yields performance similar to IPS. In contrast, the $f$-GAN based methods do not.}
    \label{fig:toy_simple}
\end{figure}
The performance of the models is measured using the mean expected reward (EXP) over all context vectors, as calculated by OBP.
Only the $f$-GAN step of the two step optimization was used, i.e.\ the divergence minimization step. 
The target policy should be similar to the logging policy after training when the divergence is the minimization objective; performance should be comparable.
The results are displayed in Figure \ref{fig:toy_divergence}.

Even though the scores for the policies trained using the directly estimated divergence $\hat d (\pi_\theta || \pi_0)$ differ significantly from the logging policy, it can be noted that these policies seem to be able to learn the distributions of the logging policy on all datasets.
In contrast, the $f$-GAN policies are unable to match the performance of the logging policy on all datasets. 
The performance of the directly estimated divergence policy improves when increasing the dataset size, whilst this is not the case for the $f$-GAN policies; only the variance decreases, in some cases.

\subsection{Multiclass classification}
Finally, the proposed simpler alternative $\hat R_\text{direct}$ was evaluated on the synthetic dataset \texttt{Synt-25-15} to study its behaviour.
The simpler method is also compared to VRCRM and the optimization objective from the counterfactual risk minimization principle \cite{swaminathan2015crm}.
The policy architectures as in Section \ref{section:div_min} were used for this experiment as well. Results are shown in Figure \ref{fig:toy_simple}.

The performance of the VRCRM policies is significantly worse than the performance of the logging policy, whilst the simpler alternative and CRM are able to match the performance of IPS. Similar findings were reported in the context of counterfactual learning to rank~\cite{gupta-2024-practical,gupta2023safe}.

\section{Conclusion}
We reproduced the experimental setting of \citeauthor{pmlr-v80-wu18g} \cite{pmlr-v80-wu18g} and proposed an alternative method to estimate and minimize a direct approximation of the divergence $d(\pi_\theta || \pi_0)$.
However, our results are not consistent with the findings by \citeauthor{pmlr-v80-wu18g}.
In particular, the $f$-GAN-based divergence minimization method seemed unable to learn the distribution of the logging distribution, which is a major component of the VRCRM method.
In contrast, our direct approximation has much better performance.

Future work can furhter explore the apparent discrepancy between our results regarding $f$-GAN divergence minimization and that of previous work.

\vspace*{-1mm}
\begin{acks}
This research was supported by Huawei Finland, 
the Dutch Research Council (NWO), under project numbers VI.Veni.222.269, 024.004.022, NWA.1389.20.\-183, and KICH3.LTP.20.006, 
and the EU's Horizon Europe program under grant No 101070212.
All content represents the opinion of the authors, which is not necessarily shared or endorsed by their respective employers and/or sponsors.
\end{acks}

\balance
\bibliographystyle{ACM-Reference-Format}
\bibliography{main}

\appendix

\section{Implementation Details}
\label{appendix:params}
\subsection{Parameters and implementation for reproduction}

The datasets and implementation of POEM are the same as in \cite{swaminathan2015crm} (\url{https://www.cs.cornell.edu/~adith/POEM/}). 
The implementation was used with default parameters.
The training function for VRCRM is published on GitHub (\url{https://github.com/hang-wu/VRCRM}). 
The two step optimization procedure involves three Adam \cite{kingma2017adam} optimizers: one for the IPS step and two for the $f$-GAN step.
However, the available code contained a bug where information was leaking between the IPS part and the $f$-GAN part of the training process due to the reuse of the IPS optimizer during the $f$-GAN step.
Additionally, the formulation of the two step optimization procedure involves a threshold on the divergence, which was not included in the implementation.

The policies were trained using the Adam optimizer~\citep{kingma2017adam} for 100 epochs using VRCRM, and ten epochs for the $f$-GAN step, with learning rate $0.001$ for the IPS part and $0.01$ for the $f$-GAN part.
Only the learning rates and model architectures (without number of nodes in the layers) were reported in \cite{pmlr-v80-wu18g}; remaining parameters were found by tuning on the validation set (25\% of the training data) or by inspecting the provided implementation.
The VRCRM policies used 1) a generator network with two hidden layers, containing 32 and 8 nodes, using batch normalization and ReLU after each linear layer and 2) a discriminator network with one hidden layer of $\lfloor (\#\text{features} + 2 \cdot \#\text{labels}) /2 \rfloor$ nodes. 
Adding the divergence constraint to VRCRM did not improve performance on the validation data.

\subsection{Synthetic data}
The synthetic datasets were generated using OBP using a logistic reward function and temperature $\beta = 5$.
Parameters were found using validation sets.
The same discriminator network as in the original setting was used for the $f$-GAN.
The generator network contained one hidden layer consisting of 15 nodes and followed by ReLU, with softmax at the end.
The policies were trained using a 0.01 learning rate and 10k batch size for 100 epochs using Adam.

\end{document}